\documentclass{article}

\usepackage{fullpage}
\usepackage{setspace}
\usepackage{parskip}
\usepackage{titlesec}
\usepackage[section]{placeins}
\usepackage{xcolor}
\usepackage{breakcites}
\usepackage{lineno}
\usepackage{hyphenat}
\usepackage{natbib}

\setcitestyle{authoryear,open={(},close={)}}

\PassOptionsToPackage{hyphens}{url}
\usepackage[colorlinks = true,
            linkcolor = blue,
            urlcolor  = blue,
            citecolor = blue,
            anchorcolor = blue]{hyperref}
\usepackage{etoolbox}

\renewenvironment{abstract}
  {{\bfseries\noindent{\abstractname}\par\nobreak}\footnotesize}
  {\bigskip}

\titlespacing{\section}{0pt}{*3}{*1}
\titlespacing{\subsection}{0pt}{*2}{*0.5}
\titlespacing{\subsubsection}{0pt}{*1.5}{0pt}

\usepackage{authblk}

\usepackage{graphicx}
\usepackage[space]{grffile}
\usepackage{latexsym}
\usepackage{textcomp}
\usepackage{longtable}
\usepackage{tabulary}
\usepackage{booktabs,array,multirow}
\usepackage{amsfonts,amsmath,amssymb}
\providecommand\citet{\cite}
\providecommand\citep{\cite}

\newif\iflatexml\latexmlfalse

\AtBeginDocument{\DeclareGraphicsExtensions{.pdf,.PDF,.eps,.EPS,.png,.PNG,.tif,.TIF,.jpg,.JPG,.jpeg,.JPEG}}

\usepackage[utf8]{inputenc}

\usepackage{float}

\usepackage [english]{babel}
\usepackage [autostyle, english = american]{csquotes}
\MakeOuterQuote{"}

\usepackage{xcolor
}
\hypersetup{
pdftitle={Transferable Models for Bioacoustics with Human Language Supervision},
pdfkeywords={Self-supervised Bioacoustics, Contrastive Language-Audio Pretraining, Animal Audio Classification, Multimodal animal data, Passive Acoustic Monitoring},
}

\iflatexml

\else
\fi

\usepackage{arxiv}
\begin{document}

\title{Transferable Models for Bioacoustics with Human Language Supervision}
\date{}

\author[1]{David Robinson}
\author[2]{Adelaide Robinson*}%
\author[3]{Lily Akrapongpisak*\thanks{Equal Contribution}}%
\affil[1]{Independent}%
\affil[2]{University of California, Santa Barbara}%
\affil[3]{University of Queensland}%
\affil[ ]{\texttt {drhearobinson@gmail.com}}

\vspace{-1em}

\begingroup
\let\center\flushleft
\let\endcenter\endflushleft
\maketitle
\endgroup

\begin{abstract}
Passive acoustic monitoring offers a scalable, non-invasive method for tracking global biodiversity and anthropogenic impacts on species. Although deep learning has become a vital tool for processing this data, current models are inflexible, typically cover only a handful of species, and are limited by data scarcity. In this work, we propose BioLingual, a new model for bioacoustics based on contrastive language-audio pretraining. We first aggregate bioacoustic archives into a language-audio dataset, called AnimalSpeak, with over a million audio-caption pairs holding information on species, vocalization context, and animal behavior. After training on this dataset to connect language and audio representations, our model can identify over a thousand species' calls across taxa, complete bioacoustic tasks zero-shot, and retrieve animal vocalization recordings from natural text queries. When fine-tuned, BioLingual sets a new state-of-the-art on nine tasks in the Benchmark of Animal Sounds. Given its broad taxa coverage and ability to be flexibly queried in human language, we believe this model opens new paradigms in ecological monitoring and research, including free-text search on the world's acoustic monitoring archives. We release our models, dataset, and code\footnote{\href{https://github.com/david-rx/biolingual}{https://github.com/david-rx/BioLingual}}. %
\end{abstract}%

\sloppy

\section*{1. Introduction}

{\label{819862}}

{Across both taxa and geographical areas, numerous animal species have
developed an array of methods to communicate using
sound \citep{Bradbury2011}. The field of bioacoustics integrates acoustic
science with biology to investigate and interpret these sounds. This
field has wide conservation applications, including allowing for the
detection, monitoring, and behavioural analysis of species of
interest~\citep{Penar2020}. It can also offer insight into the impact of
anthropogenic pressures, such as climate change and natural resource
extraction, on the behavioral ecology of species \citep{Wrege2017, Penar2020}.
Although passive acoustic monitoring can provide a cost-efficient,
reproducible, and minimally invasive method for monitoring species
compared to other field-based methods \citep{Darras2019}, it also
produces vast amounts of data requiring processing. Prior to the
development of automatic methods,~this data had to be reviewed and
labeled manually by specialists, creating a data processing bottleneck.

Machine learning offers a promising way forward. In particular, deep
learning has seen rapid adoption in bioacoustics, primarily for
automated detection and classification of various
species and call types \citep{Stowell2022}. Most commonly, convolutional neural
networks \citep{cnn}, which have had great success in the field
of computer vision, are applied to mel-spectrogram images of audio
recordings \citep{Stowell2022}. Current architectures have been applied
with high accuracy in cases with sufficient labeled data, e.g.
in \citep{Miller2023} and \citep{Bermant_2019}, allowing rapid processing
of bioacoustic data and extraction of species of interest.

Despite recent progress, current models have key limitations. The
introduction of a machine learning benchmark to the
field \citep{BEANS} has highlighted the need for architectures
that can perform well on a broad set of challenging tasks. Current
models are typically specialists designed for detecting only a handful
of species or call types \citep{BEANS, Stowell2022}. If models could be scaled
to classify thousands of species and general audio events, it could
allow retrospective monitoring of whole ecological
communities \citep{Kahl2021} and insight into fundamental ecological
questions \citep{Ross_2023}. Further, current models are inflexible,
able only to predict the fixed set of classes they were trained on. Even
with broad species coverage, this prevents researchers from rapidly
retrieving information outside of the training label schema and limits
training data in a data-scarce field \citep{Hagiwara2023} to include only
examples assigned to a fixed set of classes.~

Self-supervised models, which learn directly from raw data, offer a path
toward solving these limitations. Self-supervised learning decreases
reliance on supervised datasets, which in bioacoustics are often
labor-intensive and costly to collect \citep{Stowell2022}. Further, in
the field of natural language processing, many contemporary
self-supervised models can be given verbal prompts to perform entirely
new tasks \citep{brown2020language, wei2022finetuned}. Known as zero-shot learning, this ability
massively increases the flexibility and usability of these models.
Self-supervised learning has recently been successfully applied in
bioacoustics \citep{Hagiwara2023, Bermant_2022}. However, unlike their counterparts in
natural language processing, bioacoustic models self-supervised on raw
audio cannot utilize language prompts. Instead, they must be retrained
for each new task without transferring an understanding of the task's
objective. Because of this, the versatility of existing self-supervised
models in bioacoustics is constrained, and none has shown zero-shot
learning on a broad range of tasks.

To address the needs of bioacoustics, we turn to recent advances in
machine learning. In an impactful shift, the flexibility of language
prompting was extended to computer vision by a model
called CLIP \citep{Radford2021}. CLIP, using a technique called
contrastive learning, learned to represent images and their captions
together in a shared vision-language vector space. In this space, paired
images and captions were brought closer together, while unpaired images
and captions were pushed farther apart. By learning directly from paired
texts and images, CLIP sidestepped data scarcity and leveraged the vast
availability of captioned images on the internet. CLIP's understanding
of language enabled flexible querying to complete an open set of tasks,
and the model was competitive on many computer vision datasets even
without using their training data. Building on CLIP's success, an
analagous model for audio called~{CLAP} \citep{Elizalde2023} was
introduced, and later scaled in CLAP-LAION \citep{CLAP-LAION}. By
prompting with language, CLAP-LAION achieved state-of-the-art results in
zero-shot classification across various audio tasks, and could retrieve
relevant audios from textual queries.

In this work, we propose the use of contrastive language-audio
pretraining for bioacoustics. By leveraging natural text as labels, we
combine large-scale bioacoustic archives into a single dataset. Using
language models, we convert database metadata and free-text field notes
into descriptive audio captions to release a new dataset called
AnimalSpeak, which contains over a million text-captioned audios
spanning over 25,000 species. Adopting the best architecture explored in
CLAP-LAION \citep{CLAP-LAION}, we train on this dataset with an audio-text contrastive
objective, and transfer our model to various bioacoustic tasks. We first
show the model, which we call BioLingual, is able to retrieve relevant
bioacoustic audios given free text queries, which is not possible with
existing, audio-only models. We further find BioLingual is able to
complete a wide range of tasks across over a thousand species
"zero-shot" without further training, a first for the field. When fine-tuned, BioLingual sets a new state-of-the-art for nine
tasks in the Benchmark of Animal Sounds. Given its broad taxa coverage,
strong performance, and grounding in human language, we discuss the
model's potential for enabling new paradigms in ecological monitoring,
including indexing the world's passive acoustic monitoring data for
free-text search.

\par\null

\section*{\texorpdfstring{{2. Methods }}{2. Methods }}

\par\null

\subsection*{2.1 Training Dataset -
AnimalSpeak}

{\label{390469}}
\begin{figure}
\centering
\includegraphics[width=1.00\columnwidth]{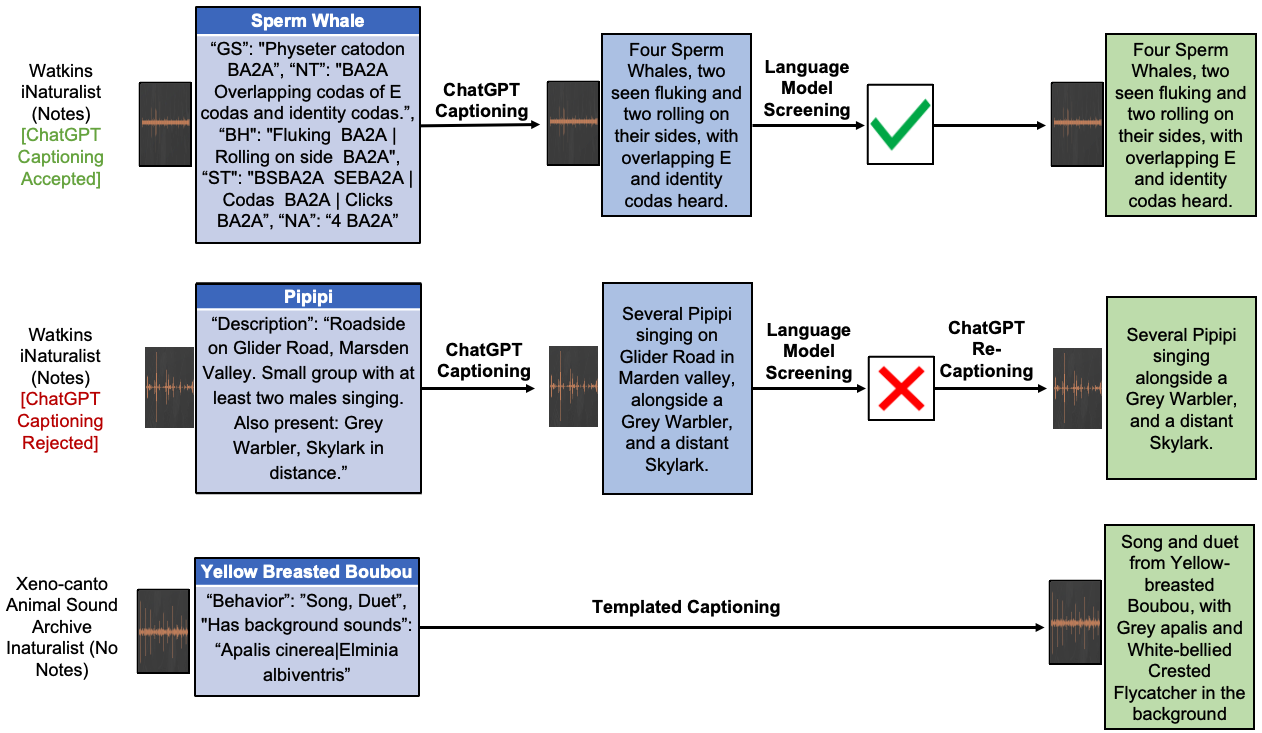}
\caption{{{Captioning AnimalSpeak language-audio Dataset. ChatGPT was used to
create create audio captions based on field notes and metadata. Pretrained
language models and regex were used to identify failure cases, which were
sent back to ChatGPT for further captioning.}
{\label{949655}}%
}}
\end{figure}

As there have been no previous works pairing human language with animal
vocalizations at scale, we began by creating an audio-to-caption dataset
for animal vocalizations. Our inputs were large-scale public collections
of animal sound recordings~including iNaturalist \citep{inaturalistSite}, Animal Sound Archive \citep{AnimalSoundArchiveAccess},
Watkins Marine Mammal Database \citep{Sayigh2016}, and Xeno-canto \citep{xenocanto}. For
each dataset, we converted recording metadata and notes into short text
captions describing the audio. We used a combination of templates,
ChatGPT \citep{openai2023chatgpt}, and postprocessing with regex and pretrained language models.
The combined dataset contains 1,102,307 text-audio pairs, covering
approximately 28,000 species. Captioned information varies greatly and
spans animal behavior, call types, descriptions of calls, number of
animals, background species, background noise and more. Captioning was
designed to remove most inaudible information, while retaining animal
behavior, relevant context, and descriptions of audible sound events.
All captions include the species of the primary animal being recorded.
The processing steps of each source dataset~{are given below~}and
depicted in Figure~{\ref{949655}}. Further examples
from each dataset are included in the appendix in
Table~{\ref{502803}}.

\emph{{iNaturalist}}{:} iNaturalist \citep{inaturalistSite} is a citizen-science platform with a
large-scale database of crowd-sourced animal observations. We included
all observations with audio files, "research-grade" species labels,
and open licensing (CC, CC-NC, and CC-NC-SA). All observations include a
species identification, and some include free-text notes left by the
recordist. The notes contain information relevant to the audio and
animal behavior, but are often long-form and mixed with non-audible
information. For recordings without notes, we used a template "The
sound of a \{species\}" as the caption. For recordings with notes, we
prompted ChatGPT to create a short audio caption summarizing audible
information, context, and behavior given the description and species
name. While many output captions were high quality, others still
contained inaudible information, particularly specific locations, and
a few did not include the species in the caption. Following
WavCaps \citep{mei2023wavcaps}, we used a pretrained named entity
recognition model, a BERT \citep{Devlin2018} model fine-tuned on
ConLL \citep{Tjong2003}, to identify all captions with specific
locations, and detected captions missing the species name by using
regex. We sent these examples back to ChatGPT with new prompts.

\emph{Xeno-Canto:~}Xeno-canto \citep{xenocanto} is a citizen-science website sharing
wildlife sounds, currently covering birds and recently
including~grasshoppers and bats. Xeno-canto recordings have extensive
metadata, including species, call type, behavior, background species,
and number of animals. We used this metadata to create a templated
caption. Xeno-canto also has free-text notes, but we leave processing
them to future work given the scale of the database and the quality of
the metadata.

\emph{Watkins Marine Mammal Database:~}The Watkins Marine Mammal
Database \citep{Sayigh2016} is an open-access collection of marine
mammal recordings collected primarily over the life of William Watkins.
We used the 15,568 annotated sound clips from the "All Cuts" release,
spanning 55 species of marine mammals. We scraped and parsed the
metadata with permission from the site curators, and extracted audible
information including signal type, number of animals, behavior, and
free-text notes. We sent this to ChatGPT with prompts to summarize with
a short audio caption. We identified common captioning failure cases
manually, and screened for these with regex. We sent flagged captions
back to ChatGPT with new prompts to remove the unnecessary information.

\emph{Animal Sound Archive:~}Animal Sound Archive \citep{AnimalSoundArchiveAccess} is a database of
animal noises maintained by the~Museum für Naturkunde Berlin. We used
the 21,474 sounds available on GBIF \citep{AnimalSoundArchive} and used the
species name with a template as the caption. To get the common names of
animal species for this dataset, we used scientific to common name
mappings from the other included datasets as well as the GBIF api.

\emph{AudioCaps:~}AudioCaps \citep{Kim_2019} is a dataset of
human-captioned audio files labeled on top of AudioSet \citep{Gemmeke2017} data. The captions are high-quality and
include many general-domain sound concepts, such as wind, human speech,
and driving cars. We are interested in developing a general approach,
and BirdNet \citep{Kahl2021} found that their detectors were most
sensitive to false positives due to noises outside the training domain.
Additionally, methods able to detect interactions of biological and
non-biological sounds have potential to answer critical ecological
questions \citep{Ross_2023}. Accordingly, we include AudioCaps to make
the method more general, improve natural language understanding, and
reduce false positives.

For all datasets besides AudioCaps, we used two captions for each audio
clip. The first caption refers to animal species by their common names, while
the second uses their scientific names. We used both under the
hypothesis that the pretrained text encoder has had more exposure to
common names during pretraining, but the scientific names are more
consistent and contain useful structure the model may learn to exploit.
All audio files were cropped or padded to ten seconds. For recordings
where the audio files were greater than ten seconds in length and
AnimalSpeak contained fewer than forty examples of the recorded species,
we chunked the audio file into a maximum of five ten-second chunks, each
with the same captions.

\par\null

\subsection*{2.2 Model Architecture, Training, and
Inference}\label{model-architecture-training-and-inference}

\begin{figure}
\centering
\includegraphics[width=1.00\columnwidth]{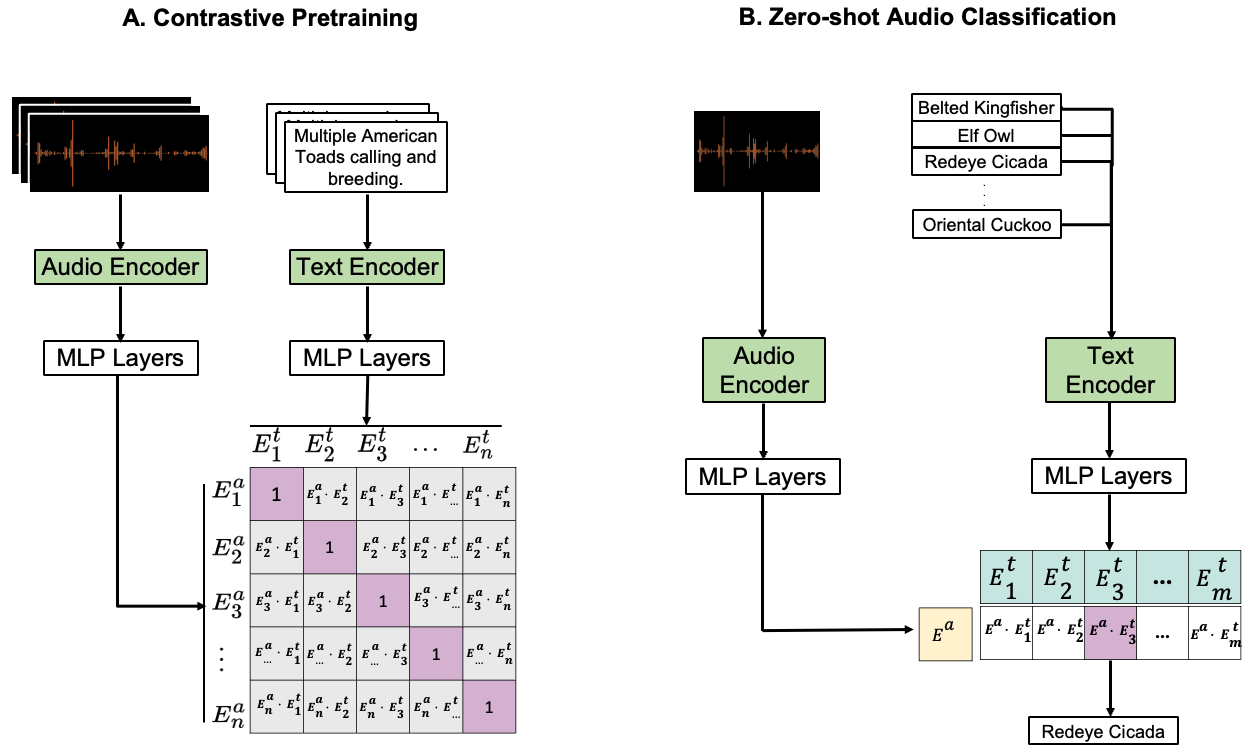}
\caption{{BioLingual model overview. BioLingual learns from paired text captions
and audios during pretraining. It performs zero-shot audio
classification by finding the most similar label representation to an audio representation.
{\label{591716}}%
}}
\end{figure}

\emph{Pretraining}

We pretrain a transformer-based \citep{Vaswani2017} language-audio model
with a contrastive objective, learning to predict which captions are
paired with which audios within a batch. Our pretraining process and model useage are depicted in Figure \ref{591716}. For the model architecture, we
reuse the best architecture investigated in
CLAP-LAION \citep{CLAP-LAION}, without feature fusion. This is an audio
encoder HTS-AT \citep{HTS-AT} operating on mel-spectrograms of
audio, and a text-encoder RoBERTa \citep{Liu2019} operating on text
captions. HTS-AT is a computationally efficient, hierarchical audio
transformer which has achieved state-of-the-art performance on several
audio classification tasks. RoBERTa is an optimization on
BERT \citep{Devlin2018}, an influential language model using only the
encoder of the original transformer architecture. A two-layer multilayer
perceptron (MLP) \citep{haykin1994neural} with ReLu
activation \citep{nair2010rectified} and output
dimension~\(D\) is added after each encoder to project the
embedding into a shared space for animal vocalizations and language.

From our training set, we sample a batch of paired captions and audios.
We obtain audio representations~\(E^a\in\ \mathbb{R}^D\)\textsubscript{~}by
passing audio~\(a\) through the audio encoder and linear
audio projection layers, and text
representations~\(E^t\in\ \mathbb{R}^D\)\textsubscript{~}by passing
text~\(t\) through the text encoder and linear text
projection layers. We use the contrastive loss proposed by
CLIP \citep{Radford2021}, bringing paired captions and audios closer in
embedding space while pushing unpaired data further apart:

\[L=\frac{1}{2N}\sum_{i=1}^N(\log\frac{\exp(\frac{E_i^a\ \cdot\ E_i^t\ }{\tau})}{\sum_{j=1}^N\exp(\frac{E_i^a\cdot\ E_j^t}{\tau})}+\log\frac{\exp(\frac{E_i^t\ \cdot\ E_i^a}{\tau})}{\sum_{j=1}^N\exp(\frac{E_i^t\ \cdot \ E_j^a}{\tau})})\]

where ~\(\tau\) is a learnable temperature parameter. This can be viewed as the average of the cross-entropy losses going from audio to text and text to audio on the prediction of true pairs within a batch. We
initialize with pretrained weights of RoBERTa and of HTS-AT-tiny,
trained respectively on general text and general audio. For examples
with multiple captions, a single caption is selected at random for each
example in each epoch.

\emph{Text-to-Audio Retrieval}

After pretraining, we use the model's cross-modal representations to
retrieve relevant audios given a text query. Specifically, given a set
of texts~\(T=t_1,t_2,...,t_m\) and a set of audios~\(A=a_1,a_2\ ...,a_n\), we
compute text embeddings~\(E^t\)\textsubscript{~~}by passing
text~\(t\) through the text encoder and linear text
projection layers, and audio embeddings~\(E^a\) by passing
audio~\(a\) through the audio encoder and linear audio
projection layers. We retrieve the most relevant audio
embeddings~\(E^a\)~for a given text embedding
\(E^t\) by ranking them according to cosine
similarity~\(S(E^t,E^a)\).~

\emph{Zero Shot Prediction}

We apply the pretrained model to unseen bioacoustic datasets or classes,
known as zero-shot prediction. To make zero-shot predictions, we create
a text prompt for each audio class label, and compute the cosine
similarity between audio embeddings and the text embeddings of the task
label prompts. Specifically, we denote the task label prompts
as~\(T=t_1, t_2, ..., t_m\) and example audios as~\(A=a_1,a_2\ ...,a_n\). We
obtain audio representations denoted
as~\(E^a\)\textsubscript{~}by passing
audio~\(a\) through the audio encoder and linear audio
projection layers, and label representations denoted
as~\(E^t\)\textsubscript{~}by passing text ~\(t\)
through the text encoder and linear text projection layers. For a given
audio example~\(a\), we compute the cosine
similarity~\(S\left(E^a,\ E^t\right)\) between the audios' embedding and each
label prompt embedding, and use these similarities as the logits for
each class. We use this framework to complete both audio classification
tasks and multilabel audio detection tasks. For classification, we take
the label whose embedding is most similar to the audio embedding. For
detection, we use the class logits as detection scores and compute
metrics such as mean-average precision by considering ordered certainty
of detection.

\emph{Fine-tuning}

In addition to zero-shot prediction, we investigate the model's ability
to transfer to different bioacoustic tasks and datasets with supervised
fine-tuning. For a given audio \(a\), we compute the audio embedding \(E^a\) by passing the audio through the audio encoder and audio projection layers, then pass this through a linear layer. For classification tasks, the output is used directly as class logits to compute a cross-entropy loss, and for detection tasks, we apply a sigmoid function before computing a binary cross-entropy loss.

\section*{3. Experiments}

{\label{541863}}

\subsection*{3.1 Pretraining and Evaluation
Setup}

{\label{884278}}

We pretrain on AnimalSpeak and apply the model to various bioacoustic
tasks and datasets. We evaluate text-to-audio retrieval, zero-shot
prediction, and fine-tuning. We include the details of evaluation datasets and
experimental setup below.

\emph{BEANS - Zero-Shot, Fine-tuning}

To best compare against prior methods, we evaluate on the Benchmark of
Animal Sounds (BEANS.) BEANS \citep{BEANS} is the first
standardized machine learning benchmark for bioacoustics, and includes
five audio detection tasks and five audio classification tasks covering
representative challenges on a diverse range of species. To evaluate
performance on general audio classification, we also include BEANS
supplementary task ESC-50 \citep{Piczak2015}. In order to exploit our
language-audio framework, we create human-readable label prompts for
all tasks in BEANS, which we release with our code. We evaluate on BEANS
zero-shot and by fine-tuning, reporting accuracy for classification
tasks and mean average precision for evaluation tasks.

\emph{Species Classification (AnimalSpeak) - Zero-Shot}

To further evaluate the model's ability to classify calls from a wide
range of species, beyond the coverage of BEANS, we evaluate species
classification on a held out AnimalSpeak test set. We randomly selected
10\% of data from the iNaturalist \citep{inaturalistSite} segment of AnimalSpeak, limited to
the 2072 species for which AnimalSpeak has at least 70 vocalizations. To
ensure independence, we included only recordings that were unique in
recording date, time and location from examples of the same species in
the training set. This sample yielded 29134 examples spanning 1143
species. We use the species names as the targets and evaluate zero-shot, using acccuracy as the metric.

\emph{AnimalSpeak - Text-to-audio retrieval}

In addition to zero-shot prediction and fine-tuning, we wish to evaluate
the model's ability to retrieve relevant audios from a text query. This
task is pertinent to the model's usability for free-text search on
acoustic monitoring data. As no other text-to-audio bioacoustic datasets
currently exist, we evaluate on the previously described independent
test set of AnimalSpeak, but using the full captions instead of species
names. Specifically, we consider each caption in the test set as a text
query, use it to retrieve audios, and evaluate whether the retrieved
audios are relevant. We use the caption containing the species common name as the query. Audios are considered relevant if their captions
are the same as the original text query after removing prefixes "The
sound of a/an " to deduplicate. We use mean average precision at ten as
our primary metric, while also reporting precision at one to help
interpret the result. Calculation details are included in the
appendix.

\emph{Pretraining Details}

BEANS overlaps with AnimalSpeak on two datasets: marine mammal
classification on Watkins \citep{Sayigh2016} and bird classification on
Cornell Bird Identification (cbi) \citep{birdsong-recognition}, where cbi is
drawn from Xeno-canto \citep{xenocanto}. To exploit AnimalSpeak without compromising
evaluation, we train two models. The first model, which we refer to as
BioLingual, is trained on the full AnimalSpeak dataset excluding the
Watkins test and validation sets from the BEANS benchmark, the cbi test
and validation sets from the BEANS benchmark, and the AnimalSpeak test set described above. The second model was trained entirely without
Watkins and Xeno-Canto. A batch size of 680 and a learning rate
of~\(1e^{-4}\) were used for the first model and a batch size
of~256 and a learning rate of ~\(7e^{-5}\) for the second.~We
pretrain both models for 45 epochs.

\par\null

\subsection*{3.2 Text-to-audio Retrieval}

\begin{table}
\centering
\includegraphics[width=0.45\columnwidth]{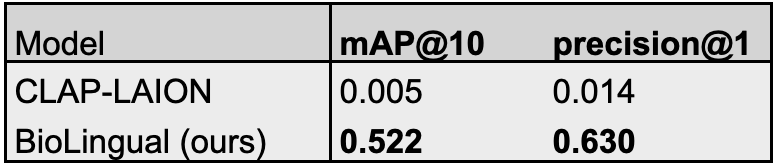}
\caption{{Text-to-audio retrieval on AnimalSpeak test set evaluates the model's ranking of
relevant audios based on their text captions. mAP@10 is mean average
precision at 10, precision@1 is precision at 1.
{\label{977719}}%
}}
\end{table}

We evaluate text-to-audio retrieval on the independent test set of
29,134 samples from AnimalSpeak, measuring the model's ability to
retrieve relevant audios from the free-text captions. We report mean
average precision at ten (mAP@10) and precision at one (precision@1),
with calculation details included in the appendix. Our results are
summarized in Table~{\ref{977719}}. Our model achieves a mAP@10 of 52.2\% and a precision@1 of 63.0\%, compared to 0.5\% and 1.4\%  from CLAP-LAION \citep{CLAP-LAION}. To our knowledge,
we are the first to report a result on text-to-audio retrieval in the field of bioacoustics, making the results difficult to contextualize. Despite being state-of-the-art on retrieval of general audio, CLAP-LAION's scores on this bioacoustic dataset are too low to serve as
a meaningful baseline. For another reference, 
we calculate the
performance of a theoretical model which retrieves audios by correctly extracting
the species from the query and recognizing the primary species with
100\% accuracy in all audios. By randomly selecting from recordings of
the correct foreground species, this theoretical model would score a
precision@1 of 64.5\% on average, representing the upper bound of a single-species detector on the 1143 species. This suggests our model, with a precision@1 of 63.0\%, nearly reaches this strong baseline by 
retrieving based on additional information in AnimalSpeak's text captions
beyond foreground species, such as background sound events, animal behaviors,
descriptions of vocalization characteristics, and the vocalization
context.

\subsection*{3.3 Zero-shot transfer}\label{zero-shot-transfer}

We evaluate our model zero-shot on BEANS \citep{BEANS}, excluding tasks involving
individual ID, and on the added AnimalSpeak species-classification test set. We compare against zero-shot
CLAP-LAION \citep{CLAP-LAION} which is state-of-the-art for multiple
zero-shot benchmarks on general audio.~{Our results are summarized in
Table~}{\ref{496646}}. Our model outperforms CLAP-LAION
on five out of the eight applicable core tasks in BEANS. Several of
these are by a large margin, with BioLingual's total core-task score at
2.2x CLAP-LAION's. Due to a lack of strong zero-shot learners in
bioacoustics, we also reference supervised models run on the BEANS
benchmark, whose results can be found in the next section in
Table~{\ref{230523}}. On cbi \citep{birdsong-recognition},
BioLingual performs remarkably well, outperforming the previous
supervised state-of-the-art model AVES \citep{Hagiwara2023}, and
indicating the model's strong performance classifying birds without
further training. This is still true even with the ablation of
Xeno-canto training, which we include in the appendix in
Table~{\ref{381688}}. On three more core tasks -
Rainforest Connection (rfcx) \citep{LEBIEN2020101113}, Eastern North American
Birds (enabirds) \citep{Chronister2021}, and Hawaiian Islands Cetacean and
Ecosystem Assessment Survey (hiceas) \citep{ncei} - zero-shot
BioLingual matches or nearly matches weaker supervised ResNet models trained from scratch
on the task. Though behind CLAP-LAION, zero-shot BioLingual outperforms
supervised ResNet models on ESC-50 \citep{Piczak2015}, indicating a
strong ability to classify general audio without further training.

We further evaluate the model's ability to classify over a large set of
species on the held-out, independent AnimalSpeak test set.~We report a
zero-shot top-1 accuracy of 68.9\% on the 1143 foreground species, while
zero-shot CLAP-LAION scores 0.4\%. To contextualize this result, the
only published model to our knowledge shown to classify at a similar
scale is BirdNet \citep{Kahl2021}, which reports a top-1 accuracy
of~77.7\% on foreground calls of 984 birds, though on a different
species and label distribution and with BirdNet fine-tuned for the
specific task.

\begin{table}
\centering
\includegraphics[width=1.00\columnwidth]{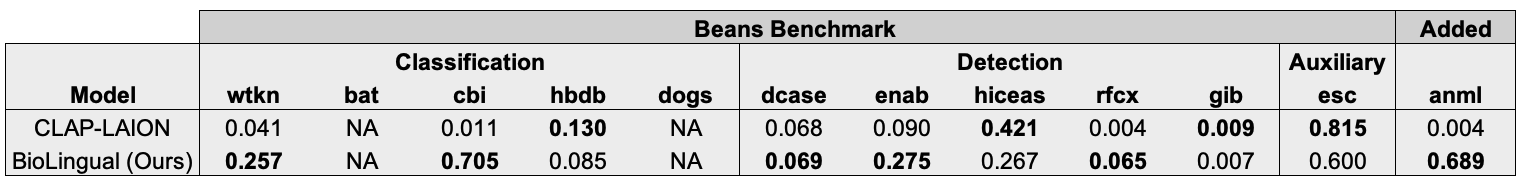}
\caption{{{Zero-shot results. Best score is bolded for each task. Added test anml is species classification on the AnimalSpeak
test set. Higher score is bolded for each task.}
{\label{496646}}%
}}
\end{table}

\subsection*{3.4 Fine-tuning - BEANS}

We evaluate our proposed model by fine-tuning on BEANS \citep{BEANS} and comparing to
ResNet \citep{He_2016_CVPR} baselines, VGGish \citep{Hershey2017}, the
current state-of-the-art model AVES \citep{Hagiwara2023}, and
CLAP-LAION \citep{CLAP-LAION}. For all experiments, we sweep the
learning rate over~\(1e^{-5}\),~\(5e^{-5}\),
and~\(1e^{-4}\), use a batch size of 32, train for 50 epochs, and
choose the best model to evaluate on the test set according to
validation set scores. Our results are summarized in Table {\ref{230523}}.

When fine-tuned, BioLingual sets a new
state-of-the-art for nine core tasks in BEANS. BioLingual also
surpasses previous bioacoustic models on ESC-50 \citep{Piczak2015}, though we exclude
CLAP-LAION from this evaluation, following the authors who also exclude
it due to data leakage in the training \citep{CLAP-LAION}. BioLingual's outperformance of CLAP-LAION corroborates the findings of \citep{ghani2023feature} by demonstrating the importance of
domain-specific pretraining data for transfer to downstream bioacoustic
tasks, as CLAP-LAION was pretrained with the same method but
approximately~\(2.5x\) more data and significantly more
compute. Based on the results of both BioLingual and CLAP-LAION, we see
a significant gain in performance from the use of models trained with contrastive language audio pretraining.

{\label{704892}}
\begin{table}
\centering
\includegraphics[width=1.00\columnwidth]{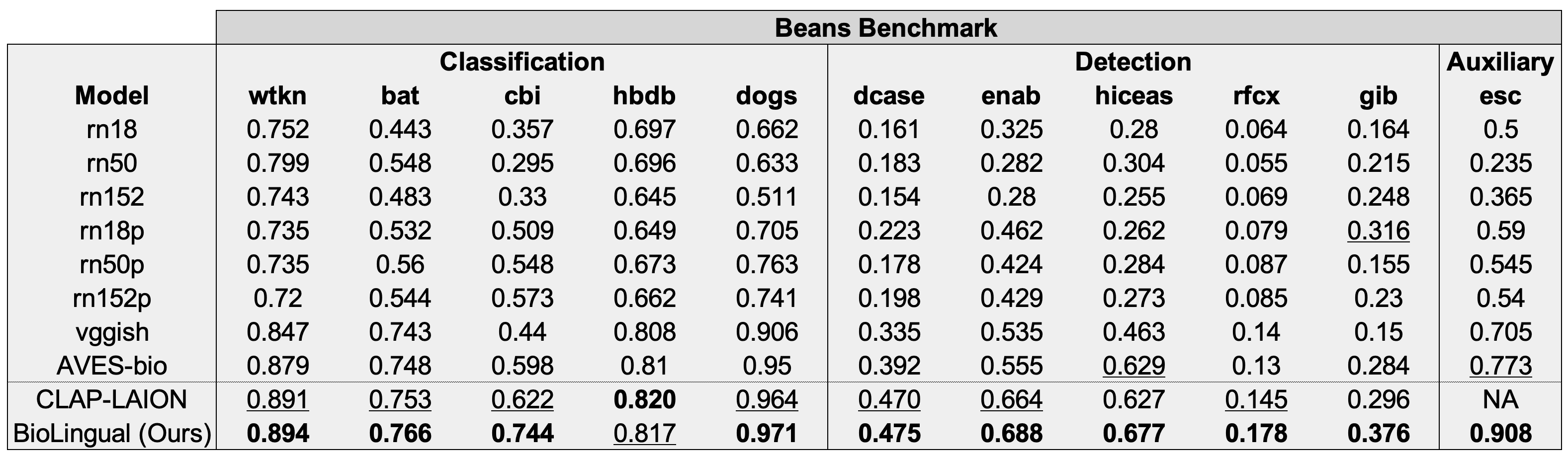}
\caption{{Fine-tuning results on BEANS. Best result is in bold and second best
underlined.
{\label{230523}}%
}}
\end{table}

\section*{\texorpdfstring{{4.
Discussion}}{4. Discussion}}

{\label{826652}}

\subsection*{4.1 Impact and Applications}

{\label{187522}}

{Having demonstrated improved performance and remarkable flexibility,
BioLingual has implications both for bioacoustics and broader ecology.
With its}~broad taxa coverage, BioLingual is able to retrospectively
analyze acoustic data to extract calls from an extensive range of
species, enabling acoustic monitoring at scale. Unlike prior methods,
BioLingual can also identify an open set of general audio events,
opening opportunities to explore relationships between anthropogenic or
natural noise and animal behavior and ecology. Separately, having
achieved state-of-the-art performance on many bioacoustic datasets when
fine-tuned, BioLingual can act as a foundation to transfer to new
bioacoustic tasks and develop detectors and classifiers with improved
performance.

As the first model capable of cross-modal text-to-audio retrieval on
bioacoustic data, we believe BioLingual opens the novel possibility of
free-text-search on passive acoustic monitoring data. With modern vector
search, this architecture supports indexing large databases such as
NOAA's passive acoustic data collection \citep{NOAA2017} or the
Australian Acoustic Observatory \citep{Roe2021}. This could be used,
for example, to quickly gain preliminary answers to ecological research
questions, to efficiently retrieve and label data for development of new
deep learning detectors, or to power citizen scientist platforms. We
note these applications merit further model evaluation, ideally
including development of a gold-standard dataset for evaluating
text-to-audio retrieval on realistic bioacoustic search queries.

\subsection*{4.2 Limitations and Future
Work}

{\label{868019}}

While natural language has proven to be a unifying interface allowing
the combination of a range of bioacoustic datasets and metadata schemas, the method is still
limited by the species distributions of the databases it draws from.
{Currently, this means an overexposure to North American and European
species and soundscapes. As the evaluation datasets have a similar bias,
this could carry over to the evaluation.~}Additionally, in part because
AnimalSpeak is primarily sourced from citizen science platforms,
important taxa such as fish are underrepresented due to the challenges
of recording these species.

We note that similar models pretrained with a contrastive
objective \citep{Radford2021, CLAP-LAION}
have used a much larger batch size, while
our model was limited in batch size by training on two 80GB GPUs. Batch
size for this model is particularly critical because the training loss
is computed by in-batch comparisons. Additionally, CLIP models have been
shown to follow well-defined scaling laws on data and
compute \citep{Cherti_2023_CVPR}. There is significant room for scale by
processing the remaining large-scale bioacoustic databases, such as the
Macaulay Library. The method can also scale naturally with the
current~proliferation of recording devices and the continued expansion
of reference libraries \citep{Parsons_2022}.

\section*{5. Conclusion}

{\label{676474}}

Passive acoustic monitoring is an increasingly adopted approach in
biological monitoring, with large archives of relatively untapped acoustic data
publicly available for analysis. Using contrastive learning on a new
language-audio dataset, we have presented a model able to transfer to
tasks across the field of bioacoustics with superior performance.
Offering the ability to{~flexibly detect~}over a thousand species and
general audio events, approaches like this can help enable
monitoring at an unprecedented scale \citep{Kahl2021}. Combined with
the novel possibility of bioacoustic text-to-audio retrieval, we hope
this work takes a step towards making the world's acoustic archives more
accessible and enabling new paradigms in ecological monitoring.

\nocite{Yin2004}
\nocite{HumBugDB}
\nocite{dcase}
\nocite{Dufourq2021}

\bibliography{references}

\begin{thebibliography}{50}
\providecommand{\natexlab}[1]{#1}
\providecommand{\url}[1]{\texttt{#1}}
\expandafter\ifx\csname urlstyle\endcsname\relax
  \providecommand{\doi}[1]{doi: #1}\else
  \providecommand{\doi}{doi: \begingroup \urlstyle{rm}\Url}\fi

\bibitem[Animal Sound Archive()]{AnimalSoundArchiveAccess}
Animal Sound Archive.
\newblock Animal sound archive, 2023.
\newblock URL
  \url{https://www.museumfuernaturkunde.berlin/en/science/animal-sound-archive}.
\newblock Accessed: May 10, 2023.

\bibitem[Bermant et~al.(2019)Bermant, Bronstein, Wood, Gero, and
  Gruber]{Bermant_2019}
Peter~C. Bermant, Michael~M. Bronstein, Robert~J. Wood, Shane Gero, and
  David~F. Gruber.
\newblock {Deep Machine Learning Techniques for the Detection and
  Classification of Sperm Whale Bioacoustics}.
\newblock \emph{Scientific Reports}, 9\penalty0 (1), Aug 2019.
\newblock \doi{10.1038/s41598-019-48909-4}.

\bibitem[Bermant et~al.(2022)Bermant, Brickson, and Titus]{Bermant_2022}
Peter~C. Bermant, Leandra Brickson, and Alexander~J. Titus.
\newblock {Bioacoustic Event Detection with Self-Supervised Contrastive
  Learning}.
\newblock Oct 2022.
\newblock \doi{10.1101/2022.10.12.511740}.

\bibitem[Bradbury and Vehrencamp(2011)]{Bradbury2011}
Jack~W. Bradbury and Sandra~L. Vehrencamp.
\newblock \emph{{Principles of animal communication, 2nd ed.}}
\newblock Vehrencamp - Oxford University Press, 2011.
\newblock ISBN 9780878930456.

\bibitem[Brown et~al.(2020)Brown, Mann, Ryder, Subbiah, Kaplan, Dhariwal,
  Neelakantan, Shyam, Sastry, Askell, et~al.]{brown2020language}
Tom Brown, Benjamin Mann, Nick Ryder, Melanie Subbiah, Jared~D Kaplan, Prafulla
  Dhariwal, Arvind Neelakantan, Pranav Shyam, Girish Sastry, Amanda Askell,
  et~al.
\newblock {Language models are few-shot learners}.
\newblock \emph{Advances in neural information processing systems},
  33:\penalty0 1877--1901, 2020.
\newblock \doi{10.48550/arXiv.2005.14165}.

\bibitem[Chen et~al.(2022)Chen, Du, Zhu, Ma, Berg-Kirkpatrick, and
  Dubnov]{HTS-AT}
Ke~Chen, Xingjian Du, Bilei Zhu, Zejun Ma, Taylor Berg-Kirkpatrick, and Shlomo
  Dubnov.
\newblock {HTS-AT: A Hierarchical Token-Semantic Audio Transformer for Sound
  Classification and Detection}.
\newblock In \emph{ICASSP 2022 - 2022 IEEE International Conference on
  Acoustics, Speech and Signal Processing (ICASSP)}, pages 646--650, 2022.
\newblock \doi{10.1109/ICASSP43922.2022.9746312}.

\bibitem[Cherti et~al.(2023)Cherti, Beaumont, Wightman, Wortsman, Ilharco,
  Gordon, Schuhmann, Schmidt, and Jitsev]{Cherti_2023_CVPR}
Mehdi Cherti, Romain Beaumont, Ross Wightman, Mitchell Wortsman, Gabriel
  Ilharco, Cade Gordon, Christoph Schuhmann, Ludwig Schmidt, and Jenia Jitsev.
\newblock {Reproducible Scaling Laws for Contrastive Language-Image Learning}.
\newblock In \emph{Proceedings of the IEEE/CVF Conference on Computer Vision
  and Pattern Recognition (CVPR)}, pages 2818--2829, Jun 2023.
\newblock \doi{10.48550/arXiv.2212.07143}.

\bibitem[Chronister et~al.(2021)Chronister, Rhinehart, Place, and
  Kitzes]{Chronister2021}
Lauren~M. Chronister, Tessa~A. Rhinehart, Aidan Place, and Justin Kitzes.
\newblock {An annotated set of audio recordings of Eastern North American birds
  containing frequency, time, and species information}.
\newblock \emph{Ecology}, 102, Jun 2021.
\newblock ISSN 19399170.
\newblock \doi{10.1002/ECY.3329}.

\bibitem[Darras et~al.(2019)Darras, Batáry, Furnas, Grass, Mulyani, and
  Tscharntke]{Darras2019}
Kevin Darras, Péter Batáry, Brett~J. Furnas, Ingo Grass, Yeni~A. Mulyani, and
  Teja Tscharntke.
\newblock {Autonomous sound recording outperforms human observation for
  sampling birds: a systematic map and user guide}.
\newblock \emph{Ecological Applications}, 29, Sep 2019.
\newblock ISSN 19395582.
\newblock \doi{10.1002/EAP.1954}.

\bibitem[Devlin et~al.(2018)Devlin, Chang, Lee, and Toutanova]{Devlin2018}
Jacob Devlin, Ming~Wei Chang, Kenton Lee, and Kristina Toutanova.
\newblock {BERT: Pre-training of Deep Bidirectional Transformers for Language
  Understanding}.
\newblock \emph{NAACL HLT 2019 - 2019 Conference of the North American Chapter
  of the Association for Computational Linguistics: Human Language Technologies
  - Proceedings of the Conference}, 1:\penalty0 4171--4186, Oct 2018.
\newblock \doi{10.18653/v1/N19-1423}.

\bibitem[Dufourq et~al.(2021)Dufourq, Durbach, Hansford, Hoepfner, Ma, Bryant,
  Stender, Li, Liu, Chen, Zhou, and Turvey]{Dufourq2021}
Emmanuel Dufourq, Ian Durbach, James~P. Hansford, Amanda Hoepfner, Heidi Ma,
  Jessica~V. Bryant, Christina~S. Stender, Wenyong Li, Zhiwei Liu, Qing Chen,
  Zhaoli Zhou, and Samuel~T. Turvey.
\newblock {Automated detection of Hainan gibbon calls for passive acoustic
  monitoring}.
\newblock \emph{Remote Sensing in Ecology and Conservation}, 7:\penalty0
  475--487, Sep 2021.
\newblock ISSN 20563485.
\newblock \doi{10.1002/RSE2.201}.

\bibitem[Elizalde et~al.(2023)Elizalde, Deshmukh, Ismail, and
  Wang]{Elizalde2023}
Benjamin Elizalde, Soham Deshmukh, Mahmoud~Al Ismail, and Huaming Wang.
\newblock {CLAP Learning Audio Concepts from Natural Language Supervision}.
\newblock \emph{ICASSP 2023 - 2023 IEEE International Conference on Acoustics,
  Speech and Signal Processing (ICASSP)}, pages 1--5, Jun 2023.
\newblock \doi{10.1109/ICASSP49357.2023.10095889}.

\bibitem[GBIF.org(2023)]{AnimalSoundArchive}
GBIF.org.
\newblock Gbif occurrence download, 2023.
\newblock URL \url{https://doi.org/10.15468/dl.w64nhg}.
\newblock Accessed: May 9 2023.

\bibitem[Gemmeke et~al.(2017)Gemmeke, Ellis, Freedman, Jansen, Lawrence, Moore,
  Plakal, and Ritter]{Gemmeke2017}
Jort~F. Gemmeke, Daniel~P.W. Ellis, Dylan Freedman, Aren Jansen, Wade Lawrence,
  R.~Channing Moore, Manoj Plakal, and Marvin Ritter.
\newblock {Audio Set: An ontology and human-labeled dataset for audio events}.
\newblock \emph{ICASSP, IEEE International Conference on Acoustics, Speech and
  Signal Processing - Proceedings}, pages 776--780, Jun 2017.
\newblock ISSN 15206149.
\newblock \doi{10.1109/ICASSP.2017.7952261}.

\bibitem[Ghani et~al.(2023)Ghani, Denton, Kahl, and Klinck]{ghani2023feature}
Burooj Ghani, Tom Denton, Stefan Kahl, and Holger Klinck.
\newblock Feature embeddings from large-scale acoustic bird classifiers enable
  few-shot transfer learning, 2023.
\newblock URL \url{https://arxiv.org/abs/2307.06292}.

\bibitem[Hagiwara(2023)]{Hagiwara2023}
Masato Hagiwara.
\newblock {AVES: Animal Vocalization Encoder Based on Self-Supervision}.
\newblock \emph{ICASSP 2023 - 2023 IEEE International Conference on Acoustics,
  Speech and Signal Processing (ICASSP)}, pages 1--5, Jun 2023.
\newblock \doi{10.1109/ICASSP49357.2023.10095642}.

\bibitem[Hagiwara et~al.(2023)Hagiwara, Hoffman, Liu, Cusimano, Effenberger,
  and Zacarian]{BEANS}
Masato Hagiwara, Benjamin Hoffman, Jen-Yu Liu, Maddie Cusimano, Felix
  Effenberger, and Katie Zacarian.
\newblock {BEANS: The Benchmark of Animal Sounds}.
\newblock In \emph{ICASSP 2023 - 2023 IEEE International Conference on
  Acoustics, Speech and Signal Processing (ICASSP)}, pages 1--5, 2023.
\newblock \doi{10.1109/ICASSP49357.2023.10096686}.

\bibitem[Haykin(1994)]{haykin1994neural}
Simon Haykin.
\newblock \emph{{Neural networks: a comprehensive foundation}}.
\newblock Prentice Hall PTR, 1994.

\bibitem[He et~al.(2016)He, Zhang, Ren, and Sun]{He_2016_CVPR}
Kaiming He, Xiangyu Zhang, Shaoqing Ren, and Jian Sun.
\newblock {Deep Residual Learning for Image Recognition}.
\newblock In \emph{Proceedings of the IEEE Conference on Computer Vision and
  Pattern Recognition (CVPR)}, Jun 2016.
\newblock \doi{10.1109/CVPR.2016.90}.

\bibitem[Hershey et~al.(2017)Hershey, Chaudhuri, Ellis, Gemmeke, Jansen, Moore,
  Plakal, Platt, Saurous, Seybold, Slaney, Weiss, and Wilson]{Hershey2017}
Shawn Hershey, Sourish Chaudhuri, Daniel P.~W. Ellis, Jort~F. Gemmeke, Aren
  Jansen, R.~Channing Moore, Manoj Plakal, Devin Platt, Rif~A. Saurous, Bryan
  Seybold, Malcolm Slaney, Ron~J. Weiss, and Kevin Wilson.
\newblock {CNN architectures for large-scale audio classification}.
\newblock In \emph{2017 IEEE International Conference on Acoustics, Speech and
  Signal Processing (ICASSP)}, pages 131--135, 2017.
\newblock \doi{10.1109/ICASSP.2017.7952132}.

\bibitem[Howard et~al.(2020)Howard, Klinck, Dane, Kahl, and tom
  denton]{birdsong-recognition}
Addison Howard, Holger Klinck, Sohier Dane, Stefan Kahl, and Tom~Denton tom
  denton.
\newblock {Cornell Birdcall Identification}, 2020.
\newblock URL \url{https://kaggle.com/competitions/birdsong-recognition}.

\bibitem[iNaturalist()]{inaturalistSite}
iNaturalist.
\newblock {iNaturalist}.
\newblock Online, 2023.
\newblock URL \url{https://www.inaturalist.org/}.
\newblock Accessed: May 1, 2023.

\bibitem[Kahl et~al.(2021)Kahl, Wood, Eibl, and Klinck]{Kahl2021}
Stefan Kahl, Connor~M. Wood, Maximilian Eibl, and Holger Klinck.
\newblock {BirdNET: A deep learning solution for avian diversity monitoring}.
\newblock \emph{Ecological Informatics}, 61:\penalty0 101236, Mar 2021.
\newblock ISSN 1574-9541.
\newblock \doi{10.1016/J.ECOINF.2021.101236}.

\bibitem[Kim et~al.(2019)Kim, Kim, Lee, and Kim]{Kim_2019}
Chris~Dongjoo Kim, Byeongchang Kim, Hyunmin Lee, and Gunhee Kim.
\newblock {AudioCaps: Generating Captions for Audios in The Wild}.
\newblock In \emph{Proceedings of the 2019 Conference of the North}.
  Association for Computational Linguistics, 2019.
\newblock \doi{10.18653/v1/n19-1011}.

\bibitem[Kiskin et~al.(2021)Kiskin, Sinka, Cobb, Rafique, Wang, Zilli,
  Gutteridge, Dam, Marinos, Li, Msaky, Kaindoa, Killeen, Herreros-Moya, Willis,
  and Roberts]{HumBugDB}
Ivan Kiskin, Marianne Sinka, Adam Cobb, Waqas Rafique, Lawrence Wang, Davide
  Zilli, Benjamin Gutteridge, Rinita Dam, Theodoros Marinos, Yunpeng Li,
  Dickson Msaky, Emmanuel Kaindoa, Gerard Killeen, Eva Herreros-Moya, Kathy
  Willis, and Stephen Roberts.
\newblock Humbugdb: A large-scale acoustic mosquito dataset, Oct 2021.
\newblock URL \url{https://arxiv.org/abs/2110.07607}.
\newblock Accessed: June, 1 2023.

\bibitem[LeBien et~al.(2020)LeBien, Zhong, Campos-Cerqueira, Velev, Dodhia,
  Ferres, and Aide]{LEBIEN2020101113}
Jack LeBien, Ming Zhong, Marconi Campos-Cerqueira, Julian~P. Velev, Rahul
  Dodhia, Juan~Lavista Ferres, and T.~Mitchell Aide.
\newblock {A pipeline for identification of bird and frog species in tropical
  soundscape recordings using a convolutional neural network}.
\newblock \emph{Ecological Informatics}, 59:\penalty0 101113, 2020.
\newblock ISSN 1574-9541.
\newblock \doi{10.1016/j.ecoinf.2020.101113}.

\bibitem[Lecun et~al.(1998)Lecun, Bottou, Bengio, and Haffner]{cnn}
Y.~Lecun, L.~Bottou, Y.~Bengio, and P.~Haffner.
\newblock {Gradient-based learning applied to document recognition}.
\newblock \emph{Proceedings of the IEEE}, 86\penalty0 (11):\penalty0
  2278--2324, 1998.
\newblock \doi{10.1109/5.726791}.

\bibitem[Liu et~al.(2019)Liu, Ott, Goyal, Du, Joshi, Chen, Levy, Lewis,
  Zettlemoyer, Stoyanov, and Allen]{Liu2019}
Yinhan Liu, Myle Ott, Naman Goyal, Jingfei Du, Mandar Joshi, Danqi Chen, Omer
  Levy, Mike Lewis, Luke Zettlemoyer, Veselin Stoyanov, and Paul~G Allen.
\newblock {RoBERTa: A Robustly Optimized BERT Pretraining Approach}.
\newblock Jul 2019.
\newblock \doi{10.48550/arXiv.1907.11692}.

\bibitem[Mei et~al.(2023)Mei, Meng, Liu, Kong, Ko, Zhao, Plumbley, Zou, and
  Wang]{mei2023wavcaps}
Xinhao Mei, Chutong Meng, Haohe Liu, Qiuqiang Kong, Tom Ko, Chengqi Zhao,
  Mark~D Plumbley, Yuexian Zou, and Wenwu Wang.
\newblock {Wavcaps: A chatgpt-assisted weakly-labelled audio captioning dataset
  for audio-language multimodal research}.
\newblock 2023.
\newblock \doi{10.48550/arXiv.2303.17395}.

\bibitem[Miller et~al.(2023)Miller, Madhusudhana, Aulich, and
  Kelly]{Miller2023}
Brian~S. Miller, Shyam Madhusudhana, Meghan~G. Aulich, and Nat Kelly.
\newblock {Deep learning algorithm outperforms experienced human observer at
  detection of blue whale D-calls: a double-observer analysis}.
\newblock \emph{Remote Sensing in Ecology and Conservation}, 9, 2023.
\newblock \doi{10.1002/rse2.297}.

\bibitem[Nair and Hinton(2010)]{nair2010rectified}
Vinod Nair and Geoffrey~E Hinton.
\newblock {Rectified linear units improve restricted boltzmann machines}.
\newblock In \emph{Proceedings of the 27th international conference on machine
  learning (ICML-10)}, pages 807--814, 2010.
\newblock \doi{10.5555/3104322.3104425}.

\bibitem[NCEI()]{ncei}
NCEI.
\newblock {Dataset Overview | National Centers for Environmental Information
  (NCEI)}.
\newblock URL \url{https://doi.org/10.25921/e12p-gj65}.
\newblock Accessed: June 1, 2023.

\bibitem[{NOAA National Centers for Environmental Information}(2017)]{NOAA2017}
{NOAA National Centers for Environmental Information}.
\newblock Passive acoustic data collection, 2017.
\newblock URL \url{https://doi.org/10.25921/PF0H-SQ72}.
\newblock Accessed: Jun 13, 2023.

\bibitem[Nolasco et~al.(2022)Nolasco, Singh, Vidaña-Vila, Grout, Morford,
  Emmerson, Jensens, Whitehead, Kiskin, Strandburg-Peshkin, Gill, Pamuła,
  Lostanlen, Morfi, and Stowell]{dcase}
Ines Nolasco, S~Singh, Ester Vidaña-Vila, E~Grout, J~Morford, M~Emmerson,
  F~Jensens, Helen Whitehead, I~Kiskin, A~Strandburg-Peshkin, Lisa Gill,
  H~Pamuła, V~Lostanlen, Veronica Morfi, and D~Stowell.
\newblock Few-shot bioacoustic sound event detection at the dcase2022
  challenge, July 2022.
\newblock URL \url{https://arxiv.org/abs/2306.09223}.

\bibitem[OpenAI(2023)]{openai2023chatgpt}
OpenAI.
\newblock Chatgpt.
\newblock \url{https://chat.openai.com/chat}, 2023.
\newblock [Large language model].

\bibitem[Parsons et~al.(2022)Parsons, Lin, Mooney, Erbe, Juanes, Lammers, Li,
  Linke, Looby, Nedelec, Opzeeland, Radford, Rice, Sayigh, Stanley, Urban, and
  Iorio]{Parsons_2022}
Miles J.~G. Parsons, Tzu-Hao Lin, T.~Aran Mooney, Christine Erbe, Francis
  Juanes, Marc Lammers, Songhai Li, Simon Linke, Audrey Looby, Sophie~L.
  Nedelec, Ilse~Van Opzeeland, Craig Radford, Aaron~N. Rice, Laela Sayigh,
  Jenni Stanley, Edward Urban, and Lucia~Di Iorio.
\newblock {Sounding the Call for a Global Library of Underwater Biological
  Sounds}.
\newblock \emph{Frontiers in Ecology and Evolution}, 10, Feb 2022.
\newblock \doi{10.3389/fevo.2022.810156}.

\bibitem[Penar et~al.(2020)Penar, Magiera, and Klocek]{Penar2020}
Weronika Penar, Angelika Magiera, and Czesław Klocek.
\newblock {Applications of bioacoustics in animal ecology}.
\newblock \emph{Ecological Complexity}, 43:\penalty0 100847, Aug 2020.
\newblock ISSN 1476-945X.
\newblock \doi{10.1016/J.ECOCOM.2020.100847}.

\bibitem[Piczak(2015)]{Piczak2015}
Karol~J. Piczak.
\newblock {ESC: Dataset for environmental sound classification}.
\newblock \emph{MM 2015 - Proceedings of the 2015 ACM Multimedia Conference},
  pages 1015--1018, Oct 2015.
\newblock \doi{10.1145/2733373.2806390}.

\bibitem[Radford et~al.(2021)Radford, Kim, Hallacy, Ramesh, Goh, Agarwal,
  Sastry, Askell, Mishkin, Clark, Krueger, and Sutskever]{Radford2021}
Alec Radford, Jong~Wook Kim, Chris Hallacy, Aditya Ramesh, Gabriel Goh,
  Sandhini Agarwal, Girish Sastry, Amanda Askell, Pamela Mishkin, Jack Clark,
  Gretchen Krueger, and Ilya Sutskever.
\newblock {Learning transferable visual models from natural language
  supervision}.
\newblock \emph{proceedings.mlr.press}, 2021.
\newblock URL \url{http://proceedings.mlr.press/v139/radford21a}.

\bibitem[Roe et~al.(2021)Roe, Eichinski, Fuller, McDonald, Schwarzkopf, Towsey,
  Truskinger, Tucker, and Watson]{Roe2021}
Paul Roe, Philip Eichinski, Richard~A. Fuller, Paul~G. McDonald, Lin
  Schwarzkopf, Michael Towsey, Anthony Truskinger, David Tucker, and David~M.
  Watson.
\newblock The australian acoustic observatory.
\newblock \emph{Methods in Ecology and Evolution}, 12\penalty0 (10):\penalty0
  1802--1808, 2021.
\newblock \doi{10.1111/2041-210X.13660}.

\bibitem[Ross et~al.(2023)Ross, O{\textquotesingle}Connell, Deichmann,
  Desjonqu{\`{e}}res, Gasc, Phillips, Sethi, Wood, and Burivalova]{Ross_2023}
Samuel R. P.-J. Ross, Darren~P. O{\textquotesingle}Connell, Jessica~L.
  Deichmann, Camille Desjonqu{\`{e}}res, Amandine Gasc, Jennifer~N. Phillips,
  Sarab~S. Sethi, Connor~M. Wood, and Zuzana Burivalova.
\newblock {Passive acoustic monitoring provides a fresh perspective on
  fundamental ecological questions}.
\newblock \emph{Functional Ecology}, 37\penalty0 (4):\penalty0 959--975, Feb
  2023.
\newblock \doi{10.1111/1365-2435.14275}.

\bibitem[Sayigh et~al.(2016)Sayigh, Daher, Allen, Gordon, Joyce, Stuhlmann, and
  Tyack]{Sayigh2016}
Laela Sayigh, Mary Daher, Julie Allen, Helen Gordon, Katherine Joyce, Claire
  Stuhlmann, and Peter Tyack.
\newblock The watkins marine mammal sound database: An online, freely
  accessible resource.
\newblock volume~27, page 040013, 1 2016.
\newblock \doi{10.1121/2.0000358}.

\bibitem[Stowell(2022)]{Stowell2022}
Dan Stowell.
\newblock {Computational bioacoustics with deep learning: a review and
  roadmap}.
\newblock \emph{PeerJ}, Mar 2022.
\newblock ISSN 21678359.
\newblock \doi{10.7717/PEERJ.13152}.

\bibitem[Tjong et~al.(2003)Tjong, Sang, and Meulder]{Tjong2003}
Erik~F Tjong, Kim Sang, and Fien~De Meulder.
\newblock {Introduction to the CoNLL-2003 Shared Task: Language-Independent
  Named Entity Recognition}, 2003.
\newblock URL \url{https://aclanthology.org/W03-0419}.

\bibitem[Vaswani et~al.(2017)Vaswani, Shazeer, Parmar, Uszkoreit, Jones, Gomez,
  Kaiser, and Polosukhin]{Vaswani2017}
Ashish Vaswani, Noam Shazeer, Niki Parmar, Jakob Uszkoreit, Llion Jones,
  Aidan~N Gomez, \L~ukasz Kaiser, and Illia Polosukhin.
\newblock {Attention is All you Need}.
\newblock In I.~Guyon, U.~Von Luxburg, S.~Bengio, H.~Wallach, R.~Fergus,
  S.~Vishwanathan, and R.~Garnett, editors, \emph{Advances in Neural
  Information Processing Systems}, volume~30. Curran Associates, Inc., 2017.
\newblock URL
  \url{https://proceedings.neurips.cc/paper_files/paper/2017/file/3f5ee243547dee91fbd053c1c4a845aa-Paper.pdf}.

\bibitem[Wei et~al.(2022)Wei, Bosma, Zhao, Guu, Yu, Lester, Du, Dai, and
  Le]{wei2022finetuned}
Jason Wei, Maarten Bosma, Vincent~Y. Zhao, Kelvin Guu, Adams~Wei Yu, Brian
  Lester, Nan Du, Andrew~M. Dai, and Quoc~V. Le.
\newblock {Finetuned Language Models Are Zero-Shot Learners}.
\newblock 2022.
\newblock \doi{10.48550/arXiv.2109.01652}.

\bibitem[Wrege et~al.(2017)Wrege, Rowland, Keen, and Shiu]{Wrege2017}
Peter~H. Wrege, Elizabeth~D. Rowland, Sara Keen, and Yu~Shiu.
\newblock {Acoustic monitoring for conservation in tropical forests: examples
  from forest elephants}.
\newblock \emph{Methods in Ecology and Evolution}, 8:\penalty0 1292--1301, Oct
  2017.
\newblock ISSN 2041210X.
\newblock \doi{10.1111/2041-210X.12730}.

\bibitem[Wu et~al.(2023)Wu, Chen, Zhang, Hui, Berg-Kirkpatrick, and
  Dubnov]{CLAP-LAION}
Yusong Wu, Ke~Chen, Tianyu Zhang, Yuchen Hui, Taylor Berg-Kirkpatrick, and
  Shlomo Dubnov.
\newblock {Large-Scale Contrastive Language-Audio Pretraining with Feature
  Fusion and Keyword-to-Caption Augmentation}.
\newblock In \emph{ICASSP 2023 - 2023 IEEE International Conference on
  Acoustics, Speech and Signal Processing (ICASSP)}, pages 1--5, 2023.
\newblock \doi{10.1109/ICASSP49357.2023.10095969}.

\bibitem[Xeno-canto()]{xenocanto}
Xeno-canto.
\newblock Xeno-canto: Bird sounds from around the world, 2023.
\newblock URL \url{https://www.xeno-canto.org/}.
\newblock Accessed: May 15, 2023.

\bibitem[Yin and McCowan(2004)]{Yin2004}
Sophia Yin and Brenda McCowan.
\newblock {Barking in domestic dogs: Context specificity and individual
  identification}.
\newblock \emph{Animal Behaviour}, 68:\penalty0 343--355, Aug 2004.
\newblock ISSN 00033472.
\newblock \doi{10.1016/J.ANBEHAV.2003.07.016}.

\end{thebibliography}

\section*{Appendix}

{\label{392625}}

\textbf{A. Details of AnimalSpeak Dataset}
\begin{table}[H]
\begin{center}
\includegraphics[width=1.00\columnwidth]{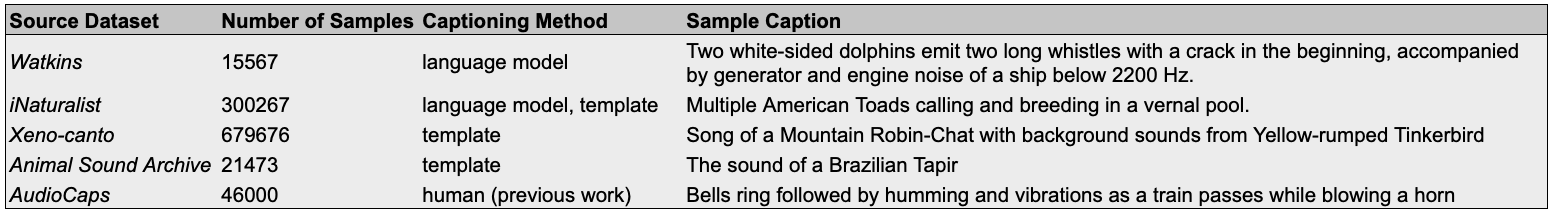}
\caption{{Contents of the AnimalSpeak Dataset
{\label{502803}}%
}}
\end{center}
\end{table}

We list the details of the~\emph{AnimalSpeak~}dataset in
Table~{\ref{502803}}, including the source datasets,
number of samples, methods used to caption, and a sample caption.

\textbf{B. Zero-shot Ablation on pretraining data}
\begin{table}[H]
\begin{center}
\includegraphics[width=1.00\columnwidth]{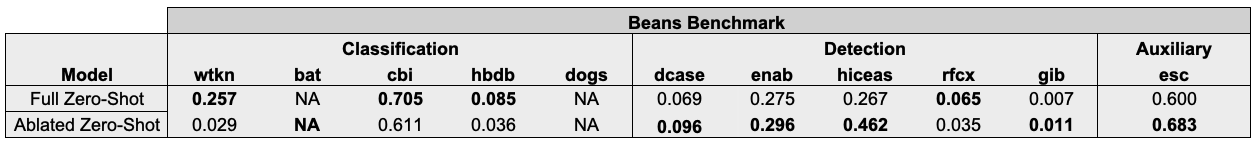}
\caption{{Zero-shot Ablation on pretraining data. Model "Ablated" was trained without
Watkins and Xeno-Canto to prevent leakage in zero-shot evaluation.~
{\label{381688}}%
}}
\end{center}
\end{table}

We compare the model trained without Xeno-canto \citep{xenocanto} and
Watkins \citep{Sayigh2016}  to the model trained on the full dataset in Table \ref{381688}.
The model trained without Xeno-canto and Watkins performs similarly in
many zero-shot tasks. The ablated model's performance on
cbi \citep{birdsong-recognition} bird classification decreases, in line with
decreasing the amount of bird recordings by more than half, but still
exceeds the previous supervised state-of-the-art. The zero-shot Watkins
marine mammal classification performance decreases dramatically, as this
smaller pretraining dataset contains very few if any hydrophone
recordings of marine mammals. While it is impossible to fully decouple
the ablation of leakage from the ablation of relevant data on Watkins,
the state-of-the-art performance on cbi demonstrates that data leakage
is not the major contributor to this model's strong zero-shot
performance on this task.

Interestingly, the ablated model performs substantially better zero-shot
on the~\emph{hiceas~}dataset \citep{ncei} detecting minke whale
calls. We hypothesize that the long-form captioning of the
\emph{Watkins} dataset in \emph{AnimalSpeak} does not lend to easy
prompting on this task.

\par\null

\textbf{C. Calculation of text-to-audio retrieval metrics}

To calculate mean average precision at~\(N\), we use the
defintion of average precision modified to account for the case of more
than~\(N\) relevant queries:

\[AP@N=\frac{1}{\min(m,N)}\sum_{k=1}^NP(k)\cdot rel(k)\]

where~\(AP@N\) is average precision at N, m is the number of
relevant items for the query,~\(P\left(k\right)\) is precision
at~\(k\), and~\(rel\left(k\right)\) is an indicator function
on the relevance of the item at ~\(k\) (1 if relevant, ~ 0 if not.)

\end{document}